%% file: emnlp2023.tex
\pdfoutput=1

\documentclass[11pt]{article}
\usepackage{amsmath}
\usepackage{EMNLP2023}
\usepackage{times}
\usepackage{latexsym}
\usepackage[T1]{fontenc}
\usepackage[utf8]{inputenc}
\usepackage{microtype}
\usepackage{inconsolata}
\usepackage{amssymb}
\usepackage{graphicx}
\usepackage{calc}
\usepackage{makecell}
\usepackage{caption}
\usepackage{subcaption}
\usepackage{cleveref}
\usepackage{color,soul}
\usepackage{xspace}
\usepackage{enumerate}
\usepackage{enumitem}
\usepackage{booktabs}
\usepackage{multirow}
\usepackage{tabularx}

\input{definitions}

\title{\pns: Instruction-Tuning LLMs for Product Title Summarization}

\author{Besnik Fetahu ~~~~ Zhiyu Chen ~~~~ Oleg Rokhlenko ~~~~ Shervin Malmasi \\
 Amazon.com, Inc. ~~~ Seattle, WA, USA \\
\texttt{\{besnikf,zhiyuche,olegro,malmasi\}@amazon.com}}

\begin{document}
\maketitle
\begin{abstract}

E-commerce product catalogs contain billions of items. Most products have lengthy titles, as sellers pack them with product attributes to improve retrieval, and highlight key product aspects.
This results in a gap between such unnatural  products titles, and how customers refer to them. It also limits how e-commerce stores can use these seller-provided titles for recommendation, QA, or review summarization.

Inspired by recent work on instruction-tuned LLMs, 
we present \pns,
a controllable approach for the task of Product Title Summarization (PTS).
Trained using a novel instruction fine-tuning strategy, our approach is able to summarize product titles according to various criteria (e.g. number of words in a summary, inclusion of specific phrases, etc.).
Extensive evaluation on a real-world e-commerce catalog shows that compared to simple fine-tuning of LLMs, our proposed approach can generate more accurate product name summaries, with an improvement of over 14 and 8 BLEU and ROUGE points, respectively.

\end{abstract}

\input{introduction}

\input{related_work}
\input{approach}
\input{setup}
\input{evaluation}
\input{conclusions}
\input{limitations}

\bibliography{anthology}
\bibliographystyle{acl_natbib}

\input{appendix}
\end{document}

%% file: definitions.tex
\usepackage{tasks}
\usepackage[many]{tcolorbox}

\definecolor{textnewGreen}{HTML}{66AE3E}
\definecolor{bgnewGreen}{HTML}{EBF4DE}
\makeatletter
\newtcolorbox{gruenebox}[1]{
  enhanced,
  skin=bicolor,
  arc=1.2mm,
  boxsep=-0.75mm,
  coltitle=white,
  colframe=textnewGreen,
  colback=bgnewGreen,
  colbacklower=white,
  fonttitle=\bfseries,
  detach title,
  title={#1},
}
\makeatother

\newcommand{\ctext}[3][RGB]{%
  \begingroup
  \definecolor{hlcolor}{#1}{#2}\sethlcolor{hlcolor}%
  \hl{#3}%
  \endgroup
}

\newcommand{\pns}{InstructPTS\xspace}
\newcommand{\instruction}[2][1=]{\ctext[RGB]{200,200,200}{\texttt{#2}\xspace}}
\newcommand{\productname}[2][1=]{{\small``\emph{#2}''\xspace}}
\newcommand{\productsummary}[2][1=]{\emph{#2}\xspace}
\newcommand{\flan}{\textsc{Flan-T5}\xspace}
\newcommand{\flanbase}{\textsc{Flan-T5-base}\xspace}
\newcommand{\flanlarge}{\textsc{Flan-T5-large}\xspace}
\newcommand{\flanxl}{\textsc{Flan-T5-xl}\xspace}
\newcommand{\type}[2][1=]{{\small{\texttt{#2}\xspace}}}
\newcommand{\summary}[2][1=]{{{\texttt{#2}}\xspace}}
\newcommand{\trainstrategy}[2][1=]{\textsc{#2}\xspace} 
\newcommand{\flancmd}[2][1=]{\textsc{Flan-T5-#2}\xspace}

%% file: introduction.tex
\section{Introduction}
\label{sec:introduction}

E-commerce product catalogs (e.g. Amazon, Walmart) contain billions of products with lengthy names: 65\% of product titles have more than 15 words~\cite{Rozen2021}.
This is due to sellers overloading titles with extra information about product functionality, colors, sizes and more in order to maximize their search rankings for as many queries as possible, and to captivate customers.  

However, this can lead to poor experiences when these titles need to be used in other contexts such as being read aloud by voice assistants, referenced in narrative text such as product summaries, or rendered in text interfaces with limited display sizes.

\begin{figure}[t]
    \centering
    \includegraphics[width=1.02\columnwidth]{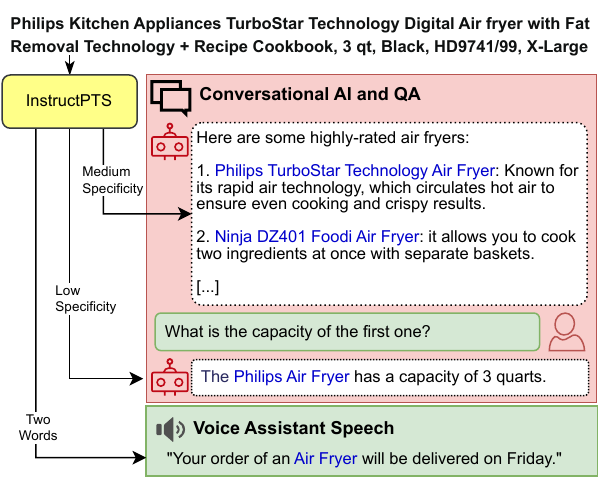}
    \caption{Example of how an original product title is reformulated by \pns for different applications.}
    \label{fig:pns_overview}
\end{figure}

This has resulted in the practical task of Product Title Summarization (PTS), which aims to extract a natural representation corresponding to how humans would refer to the product \cite{sun2018multi}.
As shown by the example in \Cref{fig:pns_overview}, these summarized titles can then be used in other tasks like voice assistant speech, product QA, summarization, recommendation, and query understanding.

Most work thus far has used traditional abstractive and extractive summarization methods to create a single summary.
Inspired by recent advances in Large Language Models (LLMs) and instruction-tuning, we present \pns, the first PTS approach
to use instruction fine-tuning (\trainstrategy{IFT}) of LLMs to achieve controllable title summarization across different dimensions such as: (i) desired length, (ii) presence of specific words (e.g. brands, size, etc.), and (iii) summary specificity.
\Cref{fig:pns_instruction_examples} shows supported instructions, which capture various requirements, and are automatically generated from a parallel dataset of original product titles and summaries. 
A key advantage of \pns is that it allows us to utilize a single model for generating multiple titles for different downstream tasks.

\begin{figure}[!ht]
\small{
\begin{gruenebox}{
\textbf{~~Example Instructions for Fine-Tuning}}
\textbf{Item Name:} ``\emph{Blade Tail Rotor Hub Set B450 330X Fusion 270 BLH1669 Replacement Helicopter Parts}''

\tcblower

 \begin{itemize}[leftmargin=*]
    \itemsep0em
     \item \texttt{Summarize \{Item\_Name\} to contain at most 3 words} $\rightarrow$ \emph{``Blade Rotor Hub''}
     \item \texttt{Summarize \{Item\_Name\} with Low specificity and to contain the words ``B450 330X''} $\rightarrow$ \emph{``Rotor Hub Set B450 330X''}
     \item \texttt{Summarize \{Item\_Name\} with Low specificity} $\rightarrow$ \emph{``Rotor Hub Set''}
 \end{itemize}
 
\end{gruenebox}}
\normalsize
\vspace{-8pt}
\caption{A sample of product title summaries generated by \pns for  different instructions. }
\label{fig:pns_instruction_examples}
\end{figure}

Evaluation on a leading real-world e-commerce catalog shows that our \pns approach generates \emph{accurate} summaries, and has high instruction-following capability.
Furthermore, the generated summaries are judged by humans as being {highly relevant} and {capturing the most salient words} from the original title.
Finally, extrinsic evaluation using a retrieval system shows that the summarized titles retain sufficient unique characteristics of the product to retrieve it with high accuracy.

%% file: related_work.tex
\section{Related Work}
\label{sec:related_work}

PTS falls within the broader domain of text summarization techniques \cite{el2021automatic}.

Both extractive and abstractive summarization approaches have been applied for PTS. For example, \citet{wang2018multi} propose a multi-task learning framework,  where one network summarizes the product name, while another learns to generate search queries. \citet{sun2018multi} propose a multi-source pointer network to generate short product names from longer input names and background knowledge. 
 \citet{gong2019automatic} developed an enhanced feature extraction approach to generate short product names by incorporating external word frequency information and named entities as additional features. 
An different approach based on Generative Adversarial Networks that encode multi-modality features (such as product images and attribute tags) is presented by \citet{zhang2019multi}.
\citet{xiao2019text} adopt Bi-LSTMs to extract key words for product name summaries.
Subsequently, \citet{mukherjee2020discriminative} tackled the vocabulary mismatch problem by integrating pre-trained embeddings with trainable character-level embeddings as inputs to Bi-LSTMs.
An adversarial generation model that can generate personalized short names is proposed by \citet{wang2020selling}. 

Our approach differs from prior work in two aspects. Firstly, previous studies primarily focused on generating a single product name summary, which may not cater to the diverse use cases in e-commerce applications.
In contrast, our approach offers the flexibility to generate diverse summary types (e.g. specific number of words, specific summary specificity etc.). 
Secondly, drawing inspiration from the recent success of LLMs \cite{ouyang2022training,longpre2023flan}, we are the first to propose an instruction-based approach for PTS.

%% file: approach.tex
\section{\pns Approach}
\label{sec:approach}

We now outline our proposed \pns approach: we describe the base model, and provide details about the instruction fine-tuning.

\subsection{Base Model}

The base model for \pns is \flan \cite{DBLP:journals/corr/abs-2210-11416}, an LLM pre-trained on a large set of instruction fine-tuning tasks.
We opt for this LLM family given that they are suitable for instruction fine-tuning (\trainstrategy{IFT}) for our task.
We experiment with different model sizes (cf. \S\ref{subsec:baselines_approach_setup}), and compare the advantage of \trainstrategy{IFT} over other training strategies.

\subsection{Ground Truth Dataset}
\label{subsec:datasets}

We use a parallel dataset of original product title and summary pairs.
The summaries are of two \emph{specificity} levels: \summary{Low} or \summary{Medium}, which control how descriptive it is w.r.t. the original title.
\summary{Low} summaries are short (approx. 2 (SD=$\pm$1) words) and typically do not include brand or other product details, but instead focus on a highly abstract description of the product family.
\summary{Medium} summaries are longer (approx. 4 (SD=$\pm$1.4) words) and contain brand/model names, and aspects that identify the specific product.
This gold data is generated using a hybrid approach: a sequence tagger chunks words that need to be included in the summary, and human annotators accept/reject the taggers decision, or rewrite the summary entirely. This is an extractive process; the summaries only contain words that appear in the original product title.

The data is split into train/dev/test sets with 100k/10k/1M product titles, respectively.
Summaries of \summary{Medium} specificity make up 58\% of the data; the remaining 42\% are of \summary{Low} specificity.
The same products can have both levels, but not always.

\subsection{Instruction Fine-Tuning}

\begin{table*}[ht]
    \centering
    \resizebox{1.0\textwidth}{!}{
\begin{tabular}{c p{12.0cm} p{4cm} p{5cm} p{5cm}}
    \toprule
    \textbf{\#} & \textbf{Instruction} & \textbf{Instruction Goal} & \textbf{Product Title (input)} & \textbf{Product Title Summary} \\
    \midrule
    1 & \instruction{Summarize \{Item Name\} with \textbf{Low} specificity} & \multirow{2}{4cm}{{Specificity Constraints.}} & \multirow{2}{5cm}{\productname{EcoSafe 6400 Certified Compostable Bags 2.5 Gallon (16" x 17"), (Case of 360 Bags : 12 Rolls)}}
    & \productsummary{Compostable Bags}\\
    2 & \instruction{Summarize \{Item Name\} with \textbf{Medium} specificity} & & & \productsummary{EcoSafe Compostable Bags}\\[3ex]
    
    \midrule 
    
    3 & \instruction{Summarize \{Item Name\} to contain at most \textbf{1} word} & \multirow{2}{4cm}{{Length Constraints.}} & \multirow{2}{5cm}{\productname{Ceramic Golden Swan/Elephant Vase Dry Flower Holder Arrangement Dining Table Home Decoration Accessories, Left Elephant}}
    & \productsummary{Vase}\\
    4 & \instruction{Summarize \{Item Name\} to contain at most \textbf{4} words} & & & \productsummary{Ceramic Golden Swan Vase}\\[5ex]
    
    \midrule
    5 & \instruction{Summarize \{Item Name\} with \textbf{Low} specificity and to contain the words \textbf{``Xbox Series S''} } & \multirow{2}{4cm}{{Phrase Inclusion Constraint.}} & \multirow{2}{5cm}{\productname{Skinit Decal Gaming Skin Compatible with Xbox Series S Controller - Officially Licensed NFL Dallas Cowboys Blast Design}}
    & \productsummary{Xbox Series S Controller Skin}\\
    6 & \instruction{Summarize \{Item Name\} with \textbf{Medium} specificity and to contain the words \textbf{``Compatible with Series S''} } & & & \productsummary{Skinit Decal Gaming Skin Compatible with Series S Controller}\\[3ex]
    
    \midrule
    7 & \instruction{Summarize \{Item Name\} by dropping up to \textbf{10} words} & \multirow{2}{4cm}{Number of deleted words constraint.} & \multirow{2}{5cm}{\productname{Girl Kayak Heartbeat Lifeline Monitor Decal Sticker 8.0 Inch BG 635}}
    & \productsummary{Decal Sticker}\\
    8 & \instruction{Summarize \{Item Name\} with \textbf{Medium} specificity and by dropping up to \textbf{5} words} & & & \productsummary{Girl Kayak Heartbeat Lifeline Monitor Decal Sticker}\\
    
    \bottomrule
\end{tabular}}
    \caption{Different instructions used by \pns to generate product title summaries.
    Each instruction has different requirements that must be satisfied in the generated summary.
    }
    \label{tab:instructions}
\end{table*}

LLM instruction fine-tuning~\citep{ouyang2022training} has proven to improve generalizability, allowing LLMs to perform better on tasks defined using natural.
\trainstrategy{IFT} allows LLMs to flexibly encode various constraints defined in natural language, enabling robust and controllable performance. 

We follow a similar approach for generating product name summaries, and fine-tune \flan models using instructions that are generated \emph{automatically} from our parallel dataset of input product names and their corresponding summaries (cf.~\S\ref{subsec:datasets}). \Cref{tab:instructions} shows the instructions used for fine-tuning \pns, as well as for generating product name summaries.

Using a product as a running example \productname{Massage Orthopedic Puzzle Floor Mat for Kids Flat Feet Prevention Sea Theme 6 Elements}, we describe in detail the instruction and the way they are constructed.

\paragraph{Specificity Level Constraints.} Instructions 1--2 in Table~\ref{tab:instructions} allow \pns to generate summaries according to the specificity levels introduced in \S\ref{subsec:datasets}. These \summary{Low} and \summary{Medium} levels allow the model to dynamically determine the summary length based on the desired specificity. Depending on the original title, the \summary{Low} specificity can yield summaries of slightly different lengths for different product.
Our training data has different levels for the same input, which helps the model learn which words are important for each specificity.

\paragraph{Word Count.} This instruction allows the model to generate summaries that contain up to a certain number of words. The instruction for training is constructed automatically, where for a product name and its ground-truth summary, depending on the number of words in the summary ($k$), we generate the instruction that has as a target the number of words equal to $k'=k+\Delta$ ($\Delta$ corresponds to a random integer $0\leq \Delta \leq 3$, where $k > 3$).  For instance, in the table below, the ground-truth summary contains 3 words, however, the instruction contains the constraint \instruction{``at most \textbf{5} words''}.
This allows the model to \emph{flexibly} use 5 words or fewer as it sees fit, 
because sometimes the most coherent summary may use fewer words due to the presence of multi-word phrases.
\vspace{-10pt}
\begin{center}
\resizebox{1.0\columnwidth}{!}{

\begin{tabular}{p{9cm}}
\toprule
\instruction{Summarize \{Item Name\} to contain at most \textbf{5} words.} $\rightarrow$ 
     \productsummary{Orthopedic Floor Mat} \\
     \bottomrule
\end{tabular}}

\end{center}

Instructions 3--4 in Table~\ref{tab:instructions} show how the same name is summarized with 1 and 4 words. The choice of words is determined automatically by the \pns model, allowing it to automatically pick the most salient words from the product name.

\paragraph{Phrase Inclusion.} In real-world settings, depending on the context, certain words may be required in the summary (e.g. brand, size, color). We automatically construct instructions from the parallel dataset by randomly choosing a word or a sequence of words from the ground-truth summary. This allows \pns to learn on how to incorporate specific phrases in the resulting summary. We evaluate the instruction following accuracy in \S\ref{sec:evaluation}.\vspace{-10pt}
\begin{center}
\resizebox{1.0\columnwidth}{!}{
\begin{tabular}{p{9cm}}
\toprule
      \instruction{Summarize \{Item Name\} with Low specificity and to contain the words ``Orthopedic''.} $\rightarrow$ \productsummary{Orthopedic Mat}\\
     \bottomrule
\end{tabular}}

\end{center}

Instructions 5--6 in \Cref{tab:instructions} show how the desired words are encoded in conjunction with categorical constraints.
This allows the model to generate summaries of different specificity, and additionally enforce the inclusion of desired phrases.

\paragraph{Deletion of $k$--words.}
Instructions 7--8 in \Cref{tab:instructions}
allow deleting up to $k$--words. This represents the reverse case of the instructions that allow the model to output summaries of specific lengths.
The instructions are inferred automatically from the ground-truth product name summary how many words need to be deleted, and additionally add a random integer  $0\leq \Delta \leq 3$. \vspace{-10pt}
\begin{center}
\resizebox{1.0\columnwidth}{!}{
\begin{tabular}{p{9cm}}
\toprule
     \instruction{Summarize \{Item Name\} by dropping up to \textbf{13} words.} $\rightarrow$ \productsummary{Orthopedic Floor Mat} \\
     \bottomrule
\end{tabular}}

\end{center}

%% file: setup.tex
\section{Experimental Setup}
\label{sec:setup}

\subsection{Evaluation Scenarios \& Metrics}
\label{subsec:eval_scenarios}

\paragraph{Automated Evaluation:} For specificity constraints, we adopt BLEU and ROUGE metrics to automatically measure summary \emph{quality} and their alignment with the ground truth. For other instructions, we compute the \emph{instruction following accuracy} of \pns, where we only assess if the model follows the constraints encoded in the instruction.\footnote{We do not assess the accuracy of the instruction for deleting $k$--words, given that this task is designed to increase model robustness rather than downstream usage. Furthermore, determining the exact number of words to be deleted to generate valid summaries is not trivial and varies across product types.}
This verifies that the summary has the desired word count, or includes a specific phrase.

\paragraph{Human and Extrinsic Evaluation:} We conduct human evaluation to assess summary quality (\S\ref{subsec:manual_eval_results}), and assess summary fidelity using retrieval (\S\ref{subsec:ir_eval_results}).

\begin{table*}[hbt!]
    \centering
    \resizebox{1.0\textwidth}{!}{
    \begin{tabular}{l l c c c c c c c c c}
    \toprule
         \textbf{Base Model} & \textbf{Strategy} & \textbf{BLEU1} & \textbf{BLEU2} & \textbf{BLEU3} & \textbf{BLEU4} & \textbf{ROUGE1} & \textbf{ROUGE2} & \textbf{ROUGE3} & \textbf{ROUGE4} & \textbf{ROUGEL}  \\
         \midrule
         \multirow{3}{*}{\flanbase} & SFT & 0.455 & 0.309 & 0.180 & 0.115 & 0.571 & 0.358 & 0.161 & 0.074 & 0.570 \\
         & CC & 0.451 & 0.307 & 0.176 & 0.114 & 0.567 & 0.356 & 0.156 & 0.073 & 0.566 \\
         & \pns & 0.585 & 0.411 & 0.247 & 0.160 & 0.665 & 0.450 & 0.230 & 0.118 & 0.663 \\[2ex]
         
         \multirow{3}{*}{\flanlarge} & SFT & 0.473 & 0.323 & 0.180 & 0.113 & 0.595 & 0.373 & 0.157 & 0.069 & 0.594\\
& CC & 0.480 & 0.331 & 0.185 & 0.117 & 0.601 & 0.382 & 0.163 & 0.073 & 0.599 \\
& \pns & 0.605 & 0.427 & 0.258 & 0.165 & 0.686 & 0.467 & 0.241 & 0.124 & 0.685 \\[2ex]

\multirow{3}{*}{\flanxl} & SFT & 0.509 & 0.356 & 0.196 & 0.120 & 0.634 & 0.408 & 0.173 & 0.075 & 0.632\\
& CC & 0.509 & 0.357 & 0.195 & 0.120 & 0.633 & 0.408 & 0.172 & 0.075 & 0.632 \\
& \pns & \textbf{0.642} & \textbf{0.463} & \textbf{0.277} & \textbf{0.173} & \textbf{0.718} & \textbf{0.502} & \textbf{0.258} & \textbf{0.127} & \textbf{0.716} \\
    \bottomrule
    \end{tabular}}
    \caption{Text generation performance as measured based on BLEU and ROUGE metrics for the different training strategies and \flan model sizes. In the case of \trainstrategy{CC} and \pns we can generate summaries according to the categorical constraints as in the ground truth (either \instruction{Low} or \instruction{Medium}), while for \trainstrategy{SFT} we can only generate a single summary, which is compared against its ground-truth counterpart (either \instruction{Low} or \instruction{Medium}).}
    \label{tab:automated_evaluation_results}
\end{table*}

\subsection{Baselines and Approach Setup}
\label{subsec:baselines_approach_setup}

We compare \pns against baselines that use different training strategies.
We also assess different \flan model sizes: (i) \flanbase, (ii) \flanlarge, and (iii) \flanxl.

\paragraph{\flancmd{SFT}:} we perform supervised fine-tuning of \flan models with input being the original product name, and the output being the ground-truth summary.
This baseline is not controllable (e.g. specificity or number of words).

\paragraph{\flancmd{CC}:} We use Control Codes (\trainstrategy{CC}) ~\cite{keskar2019ctrl} to guide summary generation. Each \trainstrategy{CC} corresponds to a specific summarization instruction, enabling controllable summarization capabilities. We use the following \trainstrategy{CC}: (i) \instruction{Low </s> \{Item Name\}}, and (ii) \instruction{Medium </s> \{Item Name\}}.

\paragraph{Training details:} please see \Cref{sec:training_details} for a detailed description of the training setup.

%% file: evaluation.tex
\section{Automatic Evaluation Results}
\label{sec:evaluation}

\Cref{tab:automated_evaluation_results} shows the automated evaluation results on the 1M title test set. We compare different \flan model sizes and the impact of the different training strategies. Output examples from \pns are shown in \Cref{sec:model_output}.

\paragraph{Text Generation Performance:}
A consistent pattern is that as model size increases, so do the BLEU and ROUGE metrics.
For instance, \flanxl improves by roughly 5 BLEU1 points  over \flanbase (for all strategies). We note a similar trend for ROUGEL.

\paragraph{Impact of Training Strategy:}
Training strategy has a significant impact. For the same model size, \pns models obtain the best performance, e.g. \pns with \flanxl obtains an improvement of 13.3 BLEU1 points over the \trainstrategy{SFT} and \trainstrategy{CC} models. Finally, we note a convergence between \trainstrategy{CC} and \trainstrategy{SFT} for the \flanxl models, with near identical performance.
Our results show the advantages of instruction tuning for PTS. 

\paragraph{Instruction Following:} 
\Cref{tab:instruction_following_accuracy} shows the instruction following accuracy for different \pns models, where we measure if the summary contains the desired number of words specified in the first instruction ({I\#1}) or includes a specific phrase as specified in the second instruction ({I\#2}) from Table~\ref{tab:instructions}.
We find that the accuracy is significantly impacted by model size. \flanxl obtains the highest instruction following accuracy among the \flan models.

\paragraph{Summary Length:}
\Cref{tab:len} shows the mean mean title length (number of words) and standard deviation for summarized titles generated for different summary types using \pns (\flanxl) on the entire test set.
For specific word counts, we find that the model generally respects the maximum length imposed in the instruction.
The categorical constraints have more variance compared to the specific word counts, and \summary{Medium} summaries have an average length of 3.80 $\pm$1.28 words.

\paragraph{Compression Ratio:}
We also analyzed the data compression ratios for \summary{Low} and \summary{Medium} summaries based on character length.
Results show high string compression ratios of 11:1 for \summary{Low} and 5:1 for \summary{Medium} summaries.
We also observed that the compression ratio varies by product category, as shown in \Cref{sec:retrieval_eval_appx}.

\begin{table}[ht!]
    \centering
    \resizebox{1\columnwidth}{!}{
    \begin{tabular}{p{2cm} p{9.2cm} l}
        \toprule
         \textbf{Model} & \textbf{Instruction} & \textbf{Acc} \\
         \midrule
         \multirow{2}{2cm}{\flanbase} & \textbf{I\#1} \instruction{Summarize \{Item Name\} to contain at most \textbf{k} words.} & 0.674\\
         & \textbf{I\#2} \instruction{Summarize \{Item Name\} to contain the words "\{T\}".} & 0.618\\[2ex]
         
         \multirow{2}{2cm}{\flanlarge} & \textbf{I\#1} \instruction{Summarize \{Item Name\} to contain at most \textbf{k} words.} & 0.673\\
         & \textbf{I\#2} \instruction{Summarize \{Item Name\} to contain the words "\{T\}".} & 0.714\\[2ex]
         
         \multirow{2}{2cm}{\flanxl} & \textbf{I\#1} \instruction{Summarize \{Item Name\} to contain at most \textbf{k} words.} & \textbf{0.765}\\
         & \textbf{I\#2} \instruction{Summarize \{Item Name\} to contain the words "\{T\}".} & \textbf{0.760}\\
    \bottomrule
    \end{tabular}}
    \caption{Instruction following accuracy for the different \pns base models using instruction fine-tuning.}
    \label{tab:instruction_following_accuracy}
\end{table}

\begin{table}[ht!]
\centering
\resizebox{0.7\columnwidth}{!}{\begin{tabular}{lcc}
\toprule
\textbf{Summary Type} & \textbf{Summary Length}  \\
\midrule
\summary{Low} & 2.07 $\pm$0.76 \\
\summary{Medium} & 3.80 $\pm$1.28 \\
\hline
\summary{1 Word} & 1.02 $\pm$0.13 \\
\summary{2 Words} & 1.95 $\pm$0.36 \\
\summary{3 Words} & 2.62 $\pm$0.63 \\
\summary{4 Words} & 3.06 $\pm$0.94 \\
\summary{5 Words} & 3.15 $\pm$1.17 \\
\bottomrule
\end{tabular}}
\caption{The mean and standard deviation of the summarized title lengths (word count) for different summary types generated by \pns (\flanxl).}
\label{tab:len}
\end{table}

\section{Human Evaluation Study}
\label{subsec:manual_eval_results}

To address the known limitations of automatic summarization evaluation, we perform a human study. 
We aim to answer the following questions:

\begin{enumerate}[label=\textbf{H\arabic*}:]
    \itemsep0em
    \item In a \emph{pairwise comparison}, which model generates better product name summaries?
    \item Are the generated summaries \emph{valid}?
    \item What is the \emph{preferred} summary length by humans for a given product name?

\end{enumerate}

\paragraph{Data}
Evaluations are carried out on a sample of 10 popular product types (e.g. \type{Electronics}). For each product type we randomly sample 10 products and generate summary titles. Detailed evaluation setup is provided in \Cref{sec:human_eval_setup_appx}.

\subsection{H1: Pairwise Summary Comparison}

We compare the two best performing models, \pns and \trainstrategy{CC} using \flanxl. For the same 100 product titles, we randomly generate either \summary{Low} or \summary{Medium} titles,\footnote{We compare only these two options, given that the \flanxl-\trainstrategy{CC} can only generate such summaries.} and ask the annotators to chose their preferred summary.
To avoid position bias, the summaries are ordered randomly.

\pns was preferred by the annotators in 55\% of the cases, while in 29\% \flanxl-\trainstrategy{CC} model was preferred. In 12\% the annotators chose \emph{both} summaries being equally good, while in 4\% of the cases, \emph{neither} title was preferred. Finally, Cohen's inter-rater agreement rate between two annotators was substantial with $\kappa=0.61$.

\subsection{H2: Validity of the Generated Summaries}
Having established that \pns generates the best summaries, two annotators judge if the summaries are valid. A summary is valid if it is \emph{coherent} and can be used to \emph{identify} at least the type of the original product. 

We generate 7 different summary types per product. \Cref{tab:hs1} shows the types and their validity scores.
On this sample of 700 titles, Cohen's inter-rater agreement was substantial ($\kappa=0.69$). 

\begin{table}[h]
    \centering
    \resizebox{.55\columnwidth}{!}{
    \begin{tabular}{l c}
    \toprule
    \textbf{Summary Type} & \textbf{Accuracy} \\
    \midrule
         \summary{Low}     & 92.5\% \\
         \summary{Medium}  & 97.5\% \\
         \hline
         \summary{1 Word}  & 39.5\% \\
         \summary{2 Words} & 78.0\% \\
         \summary{3 Words} & 85.0\% \\
         \summary{4 Words} & 90.0\% \\
         \summary{5 Words} & 96.0\% \\
         \bottomrule
    \end{tabular}}
    \caption{Validity score (binary) of the different summary types for \pns (\flanxl).}
    \label{tab:hs1}
\end{table}

The lowest scores are obtained by short summaries. The reason for that is that most products require two or more words for a summary to be meaningful w.r.t. the original product name, and be able to identify the original product. The highest scores are achieved for summaries of \summary{Medium} specificity and those with \summary{5 Words}.

\subsection{H3: Preferred Summary Length}
In this study, we aim to better understand human preferences w.r.t. summary length for the different product categories. This can help determine the summary types \pns should generate for different categories. 

Table~\ref{tab:hs2} shows the results in terms of length preferences by human annotators. We omit summaries that were deemed as not meaningful by the annotators (about 19\%).
The summaries are generated using the \pns using \flanxl model. We find a moderate agreement between annotators with a Cohen's inter-rater agreement of $\kappa=0.51$.

Across the different product categories, the preferences vary. For instance, for \type{BEAUTY}, the preferred summaries are longer, with \summary{5 words}. This is intuitive given the large variety of beauty products and brands.
On the other hand, for \type{FURNITURE}, we see that an ideal summary length is with \summary{2 words}. Such products, in most cases, can be easily summarized with few words, e.g. \productname{TV Stand}.

This study shows that ideal title summarization requires different lengths for different product categories. Our proposed \pns model can robustly summarize products of any type using either \summary{Low} or \summary{Medium} summary specificity, which have variable summary length across product categories. Additionally, we can encode various constraints in terms of phrase inclusion in the summary.
In 82\% of cases \summary{Low} summaries contain up to two words. \summary{Medium} summaries on the other hand have more than three words in 78\% of cases, with 57\% having between 3 to 4 words.
If we inspect the human preference of summary length in \Cref{tab:hs2}, we note that humans annotators tend to prefer summaries between 3--5 words, which represent summaries that have similar length as \summary{Medium} summaries.

\begin{table}[ht!]
    \centering
    \resizebox{1\columnwidth}{!}{
    \begin{tabular}{p{2cm} r r r r r}
    \toprule
    & \multicolumn{5}{c}{Preferred Length (Words)}\\
    \midrule\midrule
    Category & 1 & 2 & 3 & 4 & 5 \\
    \midrule
        \type{BOOK} & -  & - & 20\% & - & 80\%\\
        \type{SHIRT} & - & 28.6\% & 28.6\% & 14.3\% & 28.6\%\\ 
        \type{HOME} & - & 22\% & 22\% & 11\% & 44\%\\
        \type{TOY FIGURE} & - & 37.5\% & 37.5 & 37.5\%\\
        \type{SPORTING GOODS} & - & - & 62.5\% & 25\% & 12.5\%\\
        \type{BEAUTY} & - & 25\% & 12.5\% & 25\% & 37.5\% \\
        \type{TOOLS} & 12.5\% & 37.5\% & 50\% & - & -\\
        \type{FURNITURE} & - & 100\% & - & - & - \\
        \type{ELECTRONICS} & - & 33.3\% & 33.\% & 33.3\% & -\\
        \type{GROCERY} & 22\% & 67\% & 11\% & - & -\\
        \bottomrule
    \end{tabular}}
    \caption{Summary preferences across product categories. Annotators pick their preferred summaries for a sample of 10 product names per product category.
    }
    \label{tab:hs2}
\end{table}

\section{Extrinsic Evaluation with Retrieval}
\label{subsec:ir_eval_results}

We have shown that \pns can robustly summarize titles, following instructions for length and phrasal inclusion (cf. \S\ref{sec:approach}).
To assess the fidelity of the summarized titles, we perform a retrieval-based extrinsic evaluation to determine how well the original products can be retrieved by using the summary titles.
We hypothesize that a good summary with retain enough of the unique characteristics of the original product to be able to retrieve it.
Additionally, this evaluation analyzes the trade-offs between summary length vs. ranking metrics of a target product under consideration.

\paragraph{Setup:}
We use a catalog of 5M products as our testbed. 
The product titles are summarized using \pns (\flanxl) with different instructions.
The summary titles are then used as queries to review the top--$k$ products in the catalog index using the BM25 algorithm.
We also use the original title as an upper bound.

\paragraph{Evaluation:}
Evaluation is performed with standard IR metrics, Mean Reciprocal Rank (MRR) and Hit$@k$.
Higher values indicate that the summary retains more distinguishing information from the original product title.

\paragraph{Results:}
Table~\ref{t:ir} shows the ranking scores of different summary types, based on a stratified sample of 100 products from over 800 different product categories (see \Cref{sec:retrieval_eval_appx} for more details). 
Intuitively, longer summaries obtain higher ranking scores than shorter summaries, since they tend to lose more information, leading to decreased ranking accuracy. Among all instructions, \texttt{Medium} achieves the best ranking scores. As shown in Table~\ref{tab:len}, \texttt{Medium} summaries are, on average, even longer than \texttt{5 Words} summaries.

The MRR of 0.398 indicates that, on average, the ground-truth product is ranked in the 2nd and 3rd position. Furthermore, the Hit@20 score of 0.641 shows that in 64.1\% of cases the ground-truth product is featured among the top 20 results.
This study shows that our summaries retain key aspects that help identify the product in a set of 5M.
It also provides guidance on how much the titles can be compressed.

\begin{table}[!ht]
 \centering
 \resizebox{1\columnwidth}{!}{
\begin{tabular}{@{}lccc@{}}
\toprule
\textbf{Instruction} & \textbf{MRR}    & \textbf{Hit@10} & \textbf{Hit@20} \\ \midrule
Original \small{(upper bound)}                  & 0.991          & 0.998          & 0.999          \\
\bottomrule
\summary{Low}                       & 0.104          & 0.154          & 0.184          \\
\summary{Medium}                    & \textbf{0.398} & \textbf{0.566} & \textbf{0.641} \\
\summary{1 Word} & 0.008          & 0.010          & 0.016          \\
\summary{2 Words}                  & 0.104         & 0.178          & 0.225          \\
\summary{3 Words}                  & 0.220          & 0.345          & 0.416          \\
\summary{4 Words}                  & 0.281          & 0.422          & 0.487          \\
\summary{5 Words}                  & 0.286          & 0.416          & 0.480          \\
\bottomrule
\end{tabular}}
\caption{Ranking results for summaries generated by \pns (\flanxl). The first row is the upper bound, with the original product title used as a query.}\label{t:ir}
\end{table}

\section{Online Deployment}

\pns has been used in a leading global e-commerce service for various downstream shopping tasks.
It can be applied for various content generation tasks related to product summarization, comparison, question suggestion, and review summarization.
A 4k sample of generated content with embedded product titles from \pns were evaluated for quality, and 96\% were found to meet the validity criteria.

%% file: conclusions.tex
\section{Conclusion}
\label{sec:conclusions}

We presented \pns, a new approach for Product Title Summarization, and demonstrated the effectiveness of instruction-tuning for this task.
Through IFT we can train a highly accurate and controllable model for generating various types of summaries. 
Empirical studies using automatic and human evaluation studies showed that the model size has a significant impact in generating reliable and meaningful summaries, and at the same time it ensures the model's ability to follow requirements specified in the instructions.

\pns has been deployed in systems where product titles from a billion-scale catalog are summarized for various downstream applications, such as question answering and summarization.
Future work will focus on more fine-grained instructions focusing on higher levels of specificity, and support for handling constraints based on  brands/sizes/colors. 

%% file: limitations.tex
\section*{Limitations and Future Work}

Our proposed approach has some limitations that we aim to address in future work. Namely, although the generated summaries are highly meaningful and qualitative, they are constructed independently from their downstream applications. This creates a gap as to whether the most salient words for an application are chosen to be incorporated in a summary. For instance, for product retrievability, we aim at investigating whether choosing words to be incorporated in a summary can be provided by the BM25 ranking method, such that words with highest discriminative power are incorporated in the summary. We aim to do this in an end-to-end fashion, where the retrievability serves as a critic to the \pns approach providing feedback on how to change the output summary. 

Finally, we also aim to investigate the challenges in summarizing product names in conversational scenarios, where the requirements for product summaries change with every conversation turn.

%% file: appendix.tex
\clearpage
\appendix

\onecolumn

\section*{\center Appendix}

\section{Example \pns Summaries}
\label{sec:model_output}

Table~\ref{tab:example_summaries} shows example summaries generated by the \pns model using \flanxl as a base model. For each product name, 7 different summary types are generated. 

\begin{table}[!hb]
    \centering\small
    \resizebox{0.9\textwidth}{!}{
    \begin{tabular}{p{7cm} l l}
    \toprule
    \textbf{Product Title} & \textbf{Summary Type} & \textbf{Generated Summary}\\
    \midrule
      \multirow{7}{7cm}{New Balance Men's Fresh Foam Arishi V3 Classic Running Shoe, Black/Wave, 8.5}  & \summary{1 Word} & Shoe\\
 & \summary{2 Words} & Running Shoe\\
 &  \summary{3 Words} & New Balance Shoe\\
 &  \summary{4 Words} &  New Balance Running Shoe\\
 &  \summary{5 Words} &  New Balance Men's Running Shoe\\
 & \summary{Low} & Running Shoe\\
 & \summary{Medium} & New Balance Running Shoe\\
 
 \midrule
 
 \multirow{7}{7cm}{Happy Belly Frozen Chopped Kale, 12 Ounce} &  \summary{1 Word} & Kale\\
 & \summary{2 Words} & Chopped Kale\\
 & \summary{3 Words} & Happy Belly Kale\\
 & \summary{4 Words} & Happy Belly Frozen Kale\\
 & \summary{5 Words} & Happy Belly Kale\\
 & \summary{Low} & Kale\\
 & \summary{Medium} & Happy Belly Kale\\
 
 \midrule
 
  \multirow{7}{7cm}{Vinyl Wall Art Decal - Thankful - 10.5" x 23.5" - Trendy Autumn Harvest Fall Leaves Seasonal Quote for Home Bedroom Kitchen Dining Room Office Church Decoration Sticker (Orange)} &  \summary{1 Word} & Decal\\
 & \summary{2 Words} & Wall Art\\
 & \summary{3 Words} & Wall Art Decal\\
 & \summary{4 Words} & Vinyl Wall Art Decal\\
 & \summary{5 Words} & Vinyl Wall Art Decal\\
 & \summary{Low} & Wall Art Decal\\
 & \summary{Medium} & Vinyl Wall Art Decal\\
 \midrule
  \multirow{7}{7cm}{Honbay 4PCS 18mm Rubber Replacement Watch Band Strap Loops (Black)} &  \summary{1 Word} & Watch\\
 & \summary{2 Words} & Watch Band\\
 & \summary{3 Words} & Honbay Watch Band\\
 & \summary{4 Words} & Watch Band Strap Loops\\
 & \summary{5 Words} & Watch Band Strap Loops\\
 & \summary{Low} & Watch Band Strap\\
 & \summary{Medium} & Honbay Watch Band Strap\\
 \midrule
 \multirow{7}{7cm}{DECOHS 2 Packs Hanging Flower Basket Frost Cover-27.5 x 39 Inch Large Dual Drawstring Plant Protection Cover Bags-Hanging Plant Pots Frost Cover Protecting Plants from Freezing Animals Eating}  & \summary{1 Word} & Frost\\
 & \summary{2 Words} & Flower Basket\\
 &  \summary{3 Words} & DECOHS Flower Basket\\
 &  \summary{4 Words} &  DECOHS Hanging Flower Basket\\
 &  \summary{5 Words} &  DECOHS Flower Basket Frost Cover\\
 & \summary{Low} & Frost Cover\\
 & \summary{Medium} & DECOHS Flower Basket Frost Cover\\
 \midrule
    \multirow{7}{7cm}{Mens Retired Baseball Coach Shirt. Free to Do Whatever Retirement T-Shirt} & \summary{1 Word} & T-Shirt\\
 & \summary{2 Words} & Coach Shirt\\
 & \summary{3 Words} & Baseball Coach Shirt\\
 & \summary{4 Words} & Retired Baseball Coach Shirt\\
 & \summary{5 Words} & Retired Baseball Coach Shirt\\
 & \summary{Low} & T-Shirt\\
 & \summary{Medium} & Retired Baseball Coach Shirt\\
 \midrule
   \multirow{7}{7cm}{ELISORLI Compatible with Xiaomi Redmi Note 11 Pro 4G/5G Wallet Case Leather Wrist Strap Lanyard Flip Cover Card Holder Stand Phone Cases for Redme Note11 11E 11Pro Cell Accessories Women Men Black}  & \summary{1 Word} & Phone\\
 & \summary{2 Words} & Phone case\\
 &  \summary{3 Words} & ELISORLI Phone Case\\
 &  \summary{4 Words} &  ELISORLI Compatible with Xiaomi\\
 &  \summary{5 Words} &  ELISORLI Phone Case\\
 & \summary{Low} & Phone case\\
 & \summary{Medium} & ELISORLI Phone Case\\
 \midrule
 \multirow{7}{7cm}{Olive Loves Apple Promoted to Big Sister Colorful Announcement T-Shirt for Baby and Toddler Girls Sibling Outfits Chill Shirt} &  \summary{1 Word} & T-Shirt\\
 & \summary{2 Words} & Olive T-Shirt\\
 & \summary{3 Words} & Olive Loves Apple\\
 & \summary{4 Words} & Olive Loves Apple T-Shirt\\
 & \summary{5 Words} & Olive Loves Apple Promoted\\
 & \summary{Low} & T-Shirt\\
 & \summary{Medium} & Olive Loves Apple Promoted to Big Sister\\
\bottomrule
    \end{tabular}}
    \caption{Example summaries generated by the \pns model. For each product name we show 7 different summary types that are generated.}
    \label{tab:example_summaries}
\end{table}

\twocolumn

\section{Human Evaluation Setup}
\label{sec:human_eval_setup_appx}

In \S\ref{sec:evaluation} we showed the results from three human evaluation studies. The studies captured the intrinsic quality of summaries.  In H1, we compared the two best performing models to determine which summaries were preferred by human annotators. While in H2 and H3, for the best performing model, we captured validity and summary length preference by annotators.

Here we describe in detail the human evaluation setup. We carry out the annotation using two expert human annotators.
In the human evaluation studies, we focus on 10 popular e-commerce product types such as: \type{BOOK}, \type{SHIRT}, \type{HOME}, \type{TOY FIGURE}, \type{SPORTING GOOD}, \type{BEAUTY}, \type{TOOLS}, \type{FURNITURE}, \type{ELECTRONICS}, and \type{GROCERY}.

\paragraph{H1: Pairwise Summary Comparison}\mbox{}\\

For the two best performing models, \pns (\flanxl) and \flanxl-\trainstrategy{CC}, and the summary types \summary{Low} and \summary{Medium}, we compare which outputs are preferred by annotators. 

For the sample of 10 product categories, we sample randomly 10 products, and for each of the product names generate their corresponding \summary{Low} and \summary{Medium} summaries for the two models under comparison. We randomly pick either the \summary{Low} or \summary{Medium} summary from both models for the same product for comparison. This results in a total of 100 annotations by two expert annotators.

To avoid any potential position bias, we shuffle the order in which the summaries are shown the annotators, and the model information, which produces the summaries is kept hidden from the human annotators. 

An example preview of the annotation job is shown in the Table~\ref{tab:hs1_appx} below:

\begin{table}[h!]
    \centering
    \resizebox{1.0\columnwidth}{!}{
    \begin{tabular}{p{4cm} p{2.5cm} p{2.5cm} p{3cm}}
    \toprule
        \textbf{Product Name} & \textbf{Summary A} & \textbf{Summary B} & \textbf{Label}  \\
        \midrule
         \multicolumn{1}{m{4cm}}{\productname{BushKlawz Premium Prince Beard Oils Variety Set Pack Bundle of Full Size 2 oz Lumber Pacific and Urban Prince Scents and Naked Prince Scent Fragrance Set Bundle Kit}} & BushKlawz Beard Oils & Premium Beard Oils & \makecell{- Summary A\\- Summary B\\- Both\\- Neither} \\ 
         \bottomrule
    \end{tabular}}
    \caption{Annotators in this pairwise comparison choose their preferred summary, without being aware of the model that produced it. In this case summary A is generated by \pns (\flanxl), while summary B is produced by \flanxl-\trainstrategy{CC}.}
    \label{tab:hs1_appx}
\end{table}

\pagebreak
\paragraph{H2: Validity of the Generated Summaries?}\mbox{}\\

In this study, we asked the human annotators to judge whether a summary is meaningful. We defined meaningfulness as a summary which is \emph{coherent}, it can be used to \emph{identify} the product or the product type/family. 

We analyzed only the summaries generated by \pns with \flanxl as established through automated metrics, as well as the human evaluation in H1. We asked two human annotators to judge the meaningfulness of the summaries for 100 products (10 random products from 10 product categories), which resulted in a total of 700 summaries (each product name is summarized using 7 different summary types).

To judge the meaningfulness score, the annotators are shown the summary along with the original product name for judgement. Table~\ref{tab:hs2_appx} shows an example of the annotation task.

\begin{table}[ht!]
    \centering
    \resizebox{1.0\columnwidth}{!}{
\begin{tabular}{p{4cm} p{2cm} p{3.5cm} p{2.5cm}}
\toprule
\textbf{Product Title} & \textbf{Type} & \textbf{Summary} & \textbf{Is Meaningful?} \\
\midrule
\multirow{7}{4cm}{\productname{Fresh Products Bio Conqueror 105 Enzymatic Odor Counteractant Concentrate FRS 12-32BWB-MG}} & \summary{Low} & Odor Counteractant Concentrate & \makecell{Yes\\No} \\
\cline{2-4} %
& \summary{Medium} & Fresh Products Odor Counteractant Concentrate & \makecell{Yes\\No} \\
\cline{2-4} %
& \summary{1 Word} & Odor & \makecell{Yes\\No} \\
\cline{2-4} %
& \summary{2 Words} & Odor Counteractant & \makecell{Yes\\No} \\
\cline{2-4} %
& \summary{3 Words} & Fresh Products Odor & \makecell{Yes\\No} \\
\cline{2-4} %
& \summary{4 Words} & Odor Counteractant Concentrate & \makecell{Yes\\No} \\
\cline{2-4} %
& \summary{5 Words} & Fresh Enzymatic Odor Counteractant Concentrate & \makecell{Yes\\No} \\
\bottomrule
\end{tabular}}
    \caption{Annotators judge for each summary type for the given product, if the resulting summary is meaningful.}
    \label{tab:hs2_appx}
\end{table}

\paragraph{H3: Preferred Summary Length}\mbox{}\\

In this study, we gather the preference of human annotators in terms of summary length. Here too as in the previous studies, we sample 10 products from 10 product categories, and ask two human annotators to provide their preferred summary for a given product, among the 7 different summary types. Here too, the study only analyzes the summaries generated by \pns with \flanxl, given that only this model can support the flexibly generation of different summary types. Example of the annotation task is shown in Table~\ref{tab:hs3_appx}.

\begin{table}[!ht]
    \centering
    \resizebox{1.0\columnwidth}{!}{
    \begin{tabular}{p{3cm} p{6cm} p{2cm}}
    \toprule
        \textbf{Product Name} & \textbf{Summaries} & \textbf{Preferred Summary}  \\
        \midrule
        \multirow{7}{3cm}{Adidas Ultraboost 6.0 DNA X Parley Non-Dyed/Non-Dyed/Non-Dyed 8.5 D (M)} & Sneaker & \summary{1 Word}\\
 & Adidas Sneaker & \summary{2 Words}\\
 &  Adidas Running Shoe & \summary{3 Words}\\
 & Adidas Ultraboost DNA X & \summary{4 Words}\\
 &  Adidas Ultraboost DNA X Parley & \summary{5 Words}\\
 &  Running Shoe & \summary{Low}\\
 & Adidas Ultraboost DNA X Parley & \summary{Medium}\\
         \bottomrule
    \end{tabular}}
    \caption{Annotators provide their preferred summary type for a given product name, shown in the order $\{$\summary{Low}, \summary{Medium}, \summary{1 Word}, \summary{2 Words}, \summary{3 Words}, \summary{4 Words}, \summary{5 Words}$\}$.}
    \label{tab:hs3_appx}
\end{table}

\section{Retrieval Results by Product Category}
\label{sec:retrieval_eval_appx}

For extrinsic evaluation (\S \ref{subsec:ir_eval_results}), we utilized a real e-commerce product catalog, indexing a total of 5M products. To ensure an unbiased evaluation of the retrieval results presented in Table~\ref{t:ir}, we took a stratified sampling approach where 100 products were randomly selected from each product category. %
This method helped mitigate any potential biases caused by variations in the popularity of different product categories.

We selected 25 product categories and show their product-level MRR scores by \pns (\flanxl) in \Cref{t:pt_ir}, ranked by the relative decrease of MRR when transitioning from \summary{Medium} to \summary{Low} specificity:
\begin{equation}\label{eq:pt_rank}
     \frac{MRR(Medium) - MRR (Low)}{MRR(Medium)} 
\end{equation}

Additionally, to understand how much product titles are compressed, we calculate the data compression ratio (CR) of the original titles using:
\begin{equation}
    CR = \frac{len(original\ product\ title)}{len(summarized\ title)} 
\end{equation}
where the \texttt{len()} function is the string length of the titles in characters.

The results show significant variations in CRs and MRR scores across different product categories. 
Notably, product categories such as \type{BEAUTY} and \type{GROCERY} exhibit relatively lower CRs and the difference of CRs between \summary{Low} and \summary{Medium} is smaller compared to other product categories. This phenomenon can be attributed to the fact that the ground-truth of \summary{Low} summaries does not further delete more words compared with \summary{Medium}, since excessively deleting words from their names may render them less identifiable. 
Therefore, the ranking scores are relatively higher, compared to product categories like \type{EARRING} and \type{SHIRT}, whose CRs of \summary{Low} specificity can be up to 18.

\begin{table*}[t!]
\centering
\resizebox{0.8\textwidth}{!}{
\begin{tabular}{@{}lcccc@{}}
\toprule
\multicolumn{1}{c}{\textbf{Product Category}} & \textbf{MRR (Low)} & \textbf{MRR (Medium)} & \textbf{CR (Low)} & \textbf{CR (Medium)} \\ \midrule
\type{SHIRT}                                     & 0.000              & 0.280                 & 12.841            & 4.651                \\
\type{EARRING}                                   & 0.001              & 0.288                 & 16.021            & 6.620                \\
\type{NECKLACE}                                  & 0.002              & 0.322                 & 14.725            & 6.276                \\
\type{CELLULAR PHONE}                            & 0.025              & 0.318                 & 15.849            & 5.834                \\
\type{RING}                                      & 0.020              & 0.234                 & 18.025            & 6.199                \\
\type{FURNITURE}                                 & 0.039              & 0.451                 & 12.193            & 6.178                \\
\type{MASSAGER}                                  & 0.052              & 0.550                 & 11.478            & 6.246                \\
\type{TEA}                                       & 0.104              & 0.735                 & 11.813            & 4.625                \\
\type{CANDLE}                                    & 0.059              & 0.393                 & 14.571            & 5.537                \\
\type{WRENCH}                                    & 0.093              & 0.544                 & 6.932             & 3.333                \\
\type{SPEAKERS}                                  & 0.091              & 0.524                 & 8.603             & 4.891                \\
\type{PAINT}                                     & 0.060              & 0.308                 & 8.770             & 3.797                \\
\type{DRIED PLANT}                               & 0.097              & 0.470                 & 11.355            & 6.642                \\
\type{HAIR EXTENSION}                            & 0.067              & 0.306                 & 10.827            & 6.106                \\
\type{TOY FIGURE}                                & 0.105              & 0.524                 & 10.808            & 4.786                \\
\type{GUITARS}                                   & 0.094              & 0.416                 & 8.193             & 4.966                \\
\type{TOOLS}                                     & 0.124              & 0.506                 & 8.089             & 4.298                \\
\type{CONSUMER ELECTRONICS}                      & 0.124              & 0.503                 & 8.577             & 5.010                \\
\type{PRINTER}                                   & 0.102              & 0.396                 & 9.980             & 5.536                \\
\type{SPORTING GOODS}                            & 0.120              & 0.446                 & 8.835             & 4.082                \\
\type{HOME}                                      & 0.150              & 0.486                 & 11.160            & 5.134                \\
\type{MEAT}                                      & 0.264              & 0.832                 & 7.588             & 2.945                \\
\type{FRUIT}                                     & 0.268              & 0.834                 & 9.297             & 3.883                \\
\type{BEAUTY}                                    & 0.202              & 0.540                 & 9.288             & 5.170                \\
\type{GROCERY}                                   & 0.299              & 0.767                 & 6.890             & 3.079                \\
\bottomrule
\end{tabular}}
\caption{MRR scores and compression ratios (CR) for different product categories. The order of product categories is determined by Eq.~\ref{eq:pt_rank} in descending order.}\label{t:pt_ir}
\end{table*}

\section{Training Details}\label{sec:training_details}

All models are trained for a maximum of 50 epochs, with an early stopping criterion of 5 epochs of non-decreasing loss on the validation set. The batch size was set to 32. 

We used AdamW~\cite{loshchilov2017decoupled} to optimize the model's parameters. The learning rate was set to $lr=2e^{-4}$, with a 10\% of steps from the first epoch used as a linear warm-up stage to find the optimal starting $lr$. 

%% file: emnlp2023.bbl
\begin{thebibliography}{14}
\expandafter\ifx\csname natexlab\endcsname\relax\def\natexlab#1{#1}\fi

\bibitem[{Chung et~al.(2022)Chung, Hou, Longpre, Zoph, Tay, Fedus, Li, Wang,
  Dehghani, Brahma, Webson, Gu, Dai, Suzgun, Chen, Chowdhery, Narang, Mishra,
  Yu, Zhao, Huang, Dai, Yu, Petrov, Chi, Dean, Devlin, Roberts, Zhou, Le, and
  Wei}]{DBLP:journals/corr/abs-2210-11416}
Hyung~Won Chung, Le~Hou, Shayne Longpre, Barret Zoph, Yi~Tay, William Fedus,
  Eric Li, Xuezhi Wang, Mostafa Dehghani, Siddhartha Brahma, Albert Webson,
  Shixiang~Shane Gu, Zhuyun Dai, Mirac Suzgun, Xinyun Chen, Aakanksha
  Chowdhery, Sharan Narang, Gaurav Mishra, Adams Yu, Vincent~Y. Zhao, Yanping
  Huang, Andrew~M. Dai, Hongkun Yu, Slav Petrov, Ed~H. Chi, Jeff Dean, Jacob
  Devlin, Adam Roberts, Denny Zhou, Quoc~V. Le, and Jason Wei. 2022.
\newblock \href {https://doi.org/10.48550/arXiv.2210.11416} {Scaling
  instruction-finetuned language models}.
\newblock \emph{CoRR}, abs/2210.11416.

\bibitem[{El-Kassas et~al.(2021)El-Kassas, Salama, Rafea, and
  Mohamed}]{el2021automatic}
Wafaa~S El-Kassas, Cherif~R Salama, Ahmed~A Rafea, and Hoda~K Mohamed. 2021.
\newblock Automatic text summarization: A comprehensive survey.
\newblock \emph{Expert systems with applications}, 165:113679.

\bibitem[{Gong et~al.(2019)Gong, Luo, Zhu, Ou, Li, and
  Duan}]{gong2019automatic}
Yu~Gong, Xusheng Luo, Kenny~Q Zhu, Wenwu Ou, Zhao Li, and Lu~Duan. 2019.
\newblock Automatic generation of chinese short product titles for mobile
  display.
\newblock In \emph{Proceedings of the AAAI Conference on Artificial
  Intelligence}, volume~33, pages 9460--9465.

\bibitem[{Keskar et~al.(2019)Keskar, McCann, Varshney, Xiong, and
  Socher}]{keskar2019ctrl}
Nitish~Shirish Keskar, Bryan McCann, Lav~R Varshney, Caiming Xiong, and Richard
  Socher. 2019.
\newblock Ctrl: A conditional transformer language model for controllable
  generation.
\newblock \emph{arXiv preprint arXiv:1909.05858}.

\bibitem[{Longpre et~al.(2023)Longpre, Hou, Vu, Webson, Chung, Tay, Zhou, Le,
  Zoph, Wei et~al.}]{longpre2023flan}
Shayne Longpre, Le~Hou, Tu~Vu, Albert Webson, Hyung~Won Chung, Yi~Tay, Denny
  Zhou, Quoc~V Le, Barret Zoph, Jason Wei, et~al. 2023.
\newblock The flan collection: Designing data and methods for effective
  instruction tuning.
\newblock \emph{arXiv preprint arXiv:2301.13688}.

\bibitem[{Loshchilov and Hutter(2017)}]{loshchilov2017decoupled}
Ilya Loshchilov and Frank Hutter. 2017.
\newblock Decoupled weight decay regularization.
\newblock \emph{arXiv preprint arXiv:1711.05101}.

\bibitem[{Mukherjee et~al.(2020)Mukherjee, Sayapaneni, and
  Subramanya}]{mukherjee2020discriminative}
Snehasish Mukherjee, Phaniram Sayapaneni, and Shankar Subramanya. 2020.
\newblock Discriminative pre-training for low resource title compression in
  conversational grocery.
\newblock \emph{arXiv preprint arXiv:2012.06943}.

\bibitem[{Ouyang et~al.(2022)Ouyang, Wu, Jiang, Almeida, Wainwright, Mishkin,
  Zhang, Agarwal, Slama, Ray et~al.}]{ouyang2022training}
Long Ouyang, Jeffrey Wu, Xu~Jiang, Diogo Almeida, Carroll Wainwright, Pamela
  Mishkin, Chong Zhang, Sandhini Agarwal, Katarina Slama, Alex Ray, et~al.
  2022.
\newblock Training language models to follow instructions with human feedback.
\newblock \emph{Advances in Neural Information Processing Systems},
  35:27730--27744.

\bibitem[{Rozen et~al.(2021)Rozen, Carmel, Mejer, Mirkis, and
  Ziser}]{Rozen2021}
Ohad Rozen, David Carmel, Avihai Mejer, Vitaly Mirkis, and Yftah Ziser. 2021.
\newblock \href
  {https://www.amazon.science/publications/answering-product-questions-by-utilizing-questions-from-other-contextually-similar-products}
  {Answering product questions by utilizing questions from other contextually
  similar products}.
\newblock In \emph{NAACL 2021}.

\bibitem[{Sun et~al.(2018)Sun, Jiang, Sun, Pei, Ou, and Wang}]{sun2018multi}
Fei Sun, Peng Jiang, Hanxiao Sun, Changhua Pei, Wenwu Ou, and Xiaobo Wang.
  2018.
\newblock Multi-source pointer network for product title summarization.
\newblock In \emph{Proceedings of the 27th ACM International Conference on
  Information and Knowledge Management}, pages 7--16.

\bibitem[{Wang et~al.(2018)Wang, Tian, Qiu, Li, Lang, Si, and
  Lan}]{wang2018multi}
Jingang Wang, Junfeng Tian, Long Qiu, Sheng Li, Jun Lang, Luo Si, and Man Lan.
  2018.
\newblock A multi-task learning approach for improving product title
  compression with user search log data.
\newblock In \emph{Proceedings of the AAAI Conference on Artificial
  Intelligence}, volume~32.

\bibitem[{Wang et~al.(2020)Wang, Zhang, Chen, Huo, and Ren}]{wang2020selling}
Manyi Wang, Tao Zhang, Qijin Chen, Chengfu Huo, and Weijun Ren. 2020.
\newblock Selling products by machine: a user-sensitive adversarial training
  method for short title generation in mobile e-commerce.
\newblock \emph{DLP-KDD}, page~9.

\bibitem[{Xiao and Munro(2019)}]{xiao2019text}
Joan Xiao and Robert Munro. 2019.
\newblock Text summarization of product titles.
\newblock In \emph{eCOM@ SIGIR}.

\bibitem[{Zhang et~al.(2019)Zhang, Zou, Li, Wan, Pan, Gong, and
  Yu}]{zhang2019multi}
Jian-Guo Zhang, Pengcheng Zou, Zhao Li, Yao Wan, Xiuming Pan, Yu~Gong, and
  Philip~S Yu. 2019.
\newblock Multi-modal generative adversarial network for short product title
  generation in mobile e-commerce.
\newblock \emph{arXiv preprint arXiv:1904.01735}.

\end{thebibliography}
